\pdfoutput=1

\documentclass[11pt]{article}

\usepackage[dvipsnames, svgnames, x11names]{xcolor} 
\usepackage[]{acl}

\usepackage{times}
\usepackage{latexsym}

\usepackage[T1]{fontenc}

\usepackage[utf8]{inputenc}

\usepackage{microtype}

\usepackage{graphicx}
\usepackage{amsmath}
\usepackage{amsthm}
\usepackage{booktabs}
\usepackage[ruled,linesnumbered]{algorithm2e}
\usepackage{multirow}
\usepackage{amssymb}

\usepackage{multirow}
\usepackage{natbib}
\usepackage{array}
\usepackage{graphicx}
\usepackage{bbm}
\usepackage{float}
\usepackage{arydshln}
\usepackage{dsfont}
\usepackage[utf8]{inputenc}
\usepackage{cleveref}
\crefname{section}{§}{§§}
\Crefname{section}{§}{§§}
\usepackage{makecell}
\usepackage{bbding}
\usepackage{amssymb}
\usepackage{pifont}
\newcommand{\cmark}{\ding{51}}%
\newcommand{\xmark}{\ding{55}}%
\title{Parallel Instance Query Network for Named Entity Recognition}

\author{
Yongliang Shen$^{1\ast}$, Xiaobin Wang$^{2}$, Zeqi Tan$^{1}$, Guangwei Xu$^{2}$,\\ 
\textbf{Pengjun Xie$^{2}$, Fei Huang$^{2}$, Weiming Lu$^{1\dagger}$, Yueting Zhuang$^{1}$}\\
 $^{1}$College of Computer Science and Technology, Zhejiang University \\
 $^{2}$DAMO Academy, Alibaba Group\\
 \texttt{\{syl, luwm\}@zju.edu.cn}\\
  \texttt{xuanjie.wxb@alibaba-inc.com}
 }

\begin{document}

\maketitle
\renewcommand{\thefootnote}{\fnsymbol{footnote}}
\footnotetext[1]{\ \ This work was conducted when Yongliang Shen was interning at Alibaba DAMO Academy.}
\footnotetext[2]{\ \ Corresponding author.}
\renewcommand{\thefootnote}{\arabic{footnote}}
\begin{abstract}

Named entity recognition (NER) is a fundamental task in natural language processing. Recent works treat named entity recognition as a reading comprehension task, constructing type-specific queries manually to extract entities. This paradigm suffers from three issues. First, type-specific queries can only extract one type of entities per inference, which is inefficient. Second, the extraction for different types of entities is isolated, ignoring the dependencies between them. Third, query construction relies on external knowledge and is difficult to apply to realistic scenarios with hundreds of entity types. To deal with them, we propose Parallel Instance Query Network (PIQN), which sets up global and learnable instance queries to extract entities from a sentence in a parallel manner. Each instance query predicts one entity, and by feeding all instance queries simultaneously, we can query all entities in parallel. Instead of being constructed from external knowledge, instance queries can learn their different query semantics during training. For training the model, we treat label assignment as a one-to-many Linear Assignment Problem (LAP) and dynamically assign gold entities to instance queries with minimal assignment cost. Experiments on both nested and flat NER datasets demonstrate that our proposed method outperforms previous state-of-the-art models\footnote{\ Our code is available at \url{https://github.com/tricktreat/piqn}.}.
\end{abstract}

\section{Introduction}

\begin{figure}[t!]
  \centering
  \includegraphics[width=\linewidth]{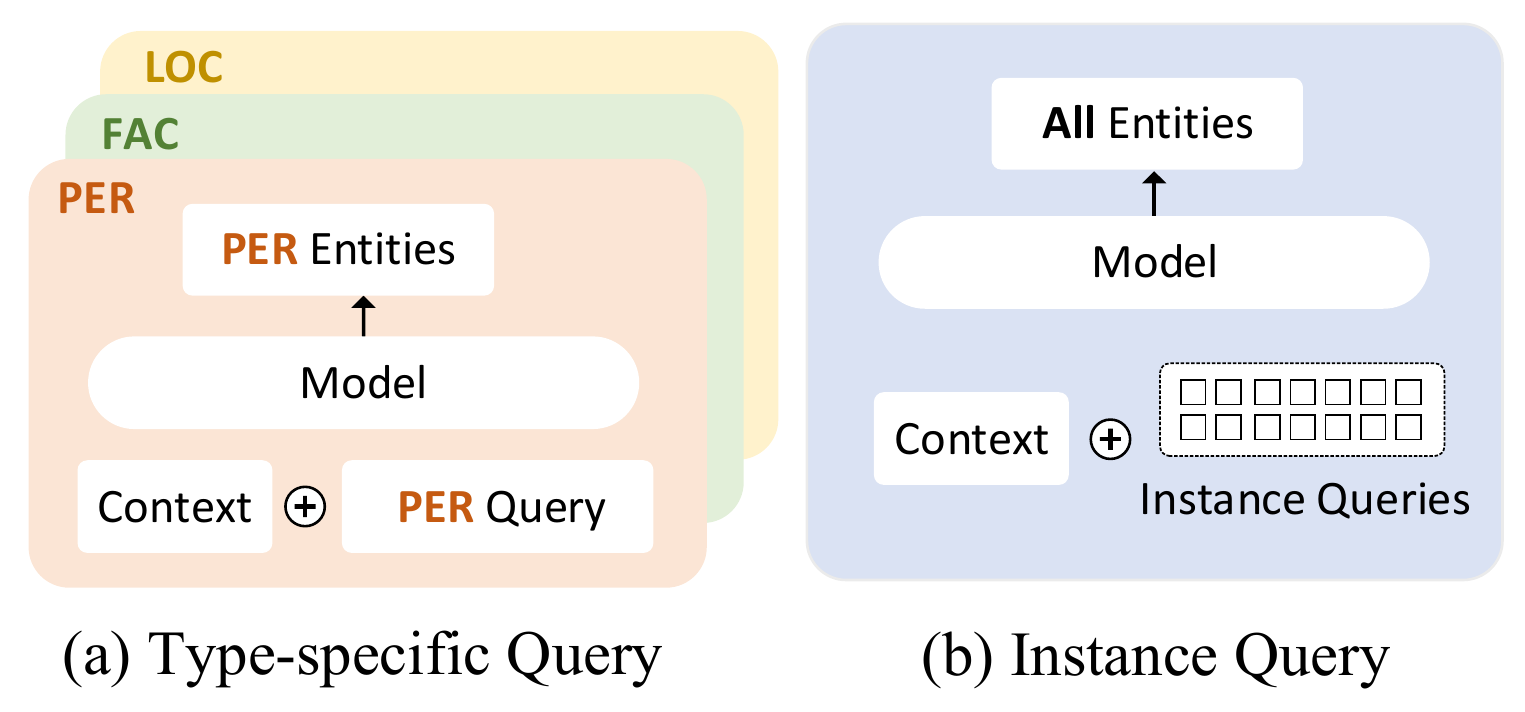}
  \caption{(a) For a sentence, type-specific queries can only extract entities of one type per inference, so the model needs to be run multiple times. (b) In contrast, instance-based queries can be input into the model simultaneously, and all entities can be extracted in parallel. Furthermore, the parallel manner can model the interactions between entities of different types.}
  \label{fig:example}
\end{figure}

Named Entity Recognition (NER) aims to identify text spans to specific entity types such as Person, Location, Organization. It has been widely used in many downstream applications such as entity linking \citep{ganea-hofmann-2017-deep, le-titov-2018-improving} and relation extraction \citep{li-ji-2014-incremental, miwa-bansal-2016-end, 10.1145/3442381.3449895}.
Traditional approaches for NER are based on sequence labeling, assigning a single tag to each word in a sentence. However, the words of nested entities have more than one tag, thus these methods lack the ability to identify nested entities.

Recently,
\citet{ ju-etal-2018-neural, strakova-etal-2019-neural, wang-etal-2020-pyramid} redesign sequence labeling models to support nested structures using different strategies.
Instead of labeling each word, \citet{luan-etal-2019-general, Tan_Qiu_Chen_Wang_Huang_2020, li-etal-2021-span, shen2021locateandlabel} perform a classification task on the text span,
and \citet{strakova-etal-2019-neural, tanl, yan2021bartner, tan2021sequencetoset} treat NER as a sequence generation or set prediction task and design encoder-decoder models to generate entities. 
Recently, 
\citet{li-etal-2020-unified, mengge-etal-2020-coarse,9463309} reformulate the NER task as a machine reading task and achieve a promising performance on both flat and nested datasets. As shown in Figure \ref{fig:example}(a), they treat the sentence as context and construct type-specific queries from external knowledge to extract entities. For example, for the sentence \textit{"U.S. President Barack Obama and his wife spent eight years in the White House"}, \citet{li-etal-2020-unified} constructs the \texttt{PER}-specific query in natural language form - \textit{"Find person entity in the text, including a single individual or a group"} to extract the \texttt{PER} entities, such as \textit{"U.S. President"}, \textit{"Barack Obama"}.
However, since the queries are type-specific, only one type of entities can be extracted for each inference.
This manner not only leads to inefficient prediction but also ignores the intrinsic connections between different types of entities, such as \textit{"U.S."} and \textit{"U.S. President"}.
In addition, type-specific queries rely on external knowledge for manual construction, which makes it difficult to fit realistic scenarios with hundreds of entity types.


In this paper, we propose the Parallel Instance Query Network (PIQN), where global and learnable instance queries replace type-specific ones to extract entities in parallel. As shown in Figure \ref{fig:example}(b), each instance query predicts one entity, and multiple instance queries can be fed simultaneously to predict all entities. Different from previous methods, we do not need external knowledge to construct the query into natural language form. The instance query can learn different query semantics during training, such as position-related or type-related semantics.
Since the semantics of instance queries are implicit, we cannot assign gold entities as their labels in advance. To tackle this, we treat label assignment as a one-to-many Linear Assignment Problem (LAP)
\citep{Burkard1999LinearAP},
and design a dynamic label assignment mechanism to assign gold entities for instance queries.

Our main contributions are as follow:

\begin{itemize}
    \item Different from type-specific queries that require multiple rounds of query, our model employs instance queries that can extract all entities in parallel. Furthermore, the style of parallel query can model the interactions between entities of different types.
    \item Instead of relying on external knowledge to construct queries in natural language form, instance queries learn their query semantics related to entity location and entity type during training.
    \item 
    To train the model,
    we design a dynamic one-to-many label assignment mechanism,
    where the entities are dynamically assigned as labels for the instance queries during training.
    The one-to-many manner allows multiple queries to predict the same entity, which can further improve the model performance.
    \item Experiments show that our model achieves state-of-the-art performance consistently on several nested and flat NER datasets.
\end{itemize}

\section{Related Work}

Traditional approaches for NER can be divided into three categories, including tagging-based, hypergraph-based and span-based approaches. The typical sequence labeling approach \citep{huang2015bidirectional} predicts labels for each token, and struggles to address nested NER. Some works \citep{alex-etal-2007-recognising, wang-etal-2020-pyramid} adapt the sequence labeling model to nested entity structures by designing a special tagging scheme. Different from the decoding on the linear sequence, the hypergraph-based approaches \citep{lu-roth-2015-joint, muis-lu-2017-labeling, katiyar-cardie-2018-nested} construct hypergraphs based on the entity nesting structure and decode entities on the hypergraph. Span-based methods first extract spans by enumeration \citep{sohrab-miwa-2018-deep, luan-etal-2019-general} or boundary identification \citep{zheng-etal-2019-boundary, Tan_Qiu_Chen_Wang_Huang_2020}, and then classify the spans. Based on these, \citet{shen2021locateandlabel} treats NER as a joint task of boundary regression and span classification and proposes a two-stage identifier of locating entities first and labeling them later.

Three novel paradigms for NER have recently been proposed, reformulating named entity recognition as sequence generation, set prediction, and reading comprehension tasks, respectively. \citet{yan2021bartner} formulates NER as an entity span sequence generation problem and uses a BART \citep{lewis-etal-2020-bart} model with the pointer mechanism to tackle  NER tasks. \citet{tan2021sequencetoset} formulates NER as an entity set prediction task. Different from \citet{strakova-etal-2019-neural}, they  utilize a non-autoregressive decoder to predict entity set. \citet{li-etal-2020-unified, mengge-etal-2020-coarse} reformulate the NER task as an MRC question answering task. They construct type-specific queries using semantic prior information for entity categories.

Different from \citet{li-etal-2020-unified, jiang2021new}, our method attempts to query at the entity level, where it adaptively learns query semantics for instance queries and extracts all types of entities in parallel. 
It is worth noting that Seq2Set \citep{tan2021sequencetoset} is quite different from ours: (1) Seq2Set attempts to eliminate the incorrect bias introduced by specified entity decoding order in the seq2seq framework, and proposes an entity set predictor, while we follow the MRC paradigm and focus on extracting entities using instance queries. (2) Seq2Set is an encoder-decoder architecture, while our model throws away the decoder and keeps only the encoder as in \citet{wenwang}, which speeds up inference and allows full interaction between query and context.
(3) Seq2Set uses bipartite graph matching to compute the entity-set level loss, while we focus on the label assignment for each instance query and propose a one-to-many dynamic label assignment mechanism.

\section{Method}

\begin{figure*}[h]
  \centering
  \includegraphics[width=\linewidth]{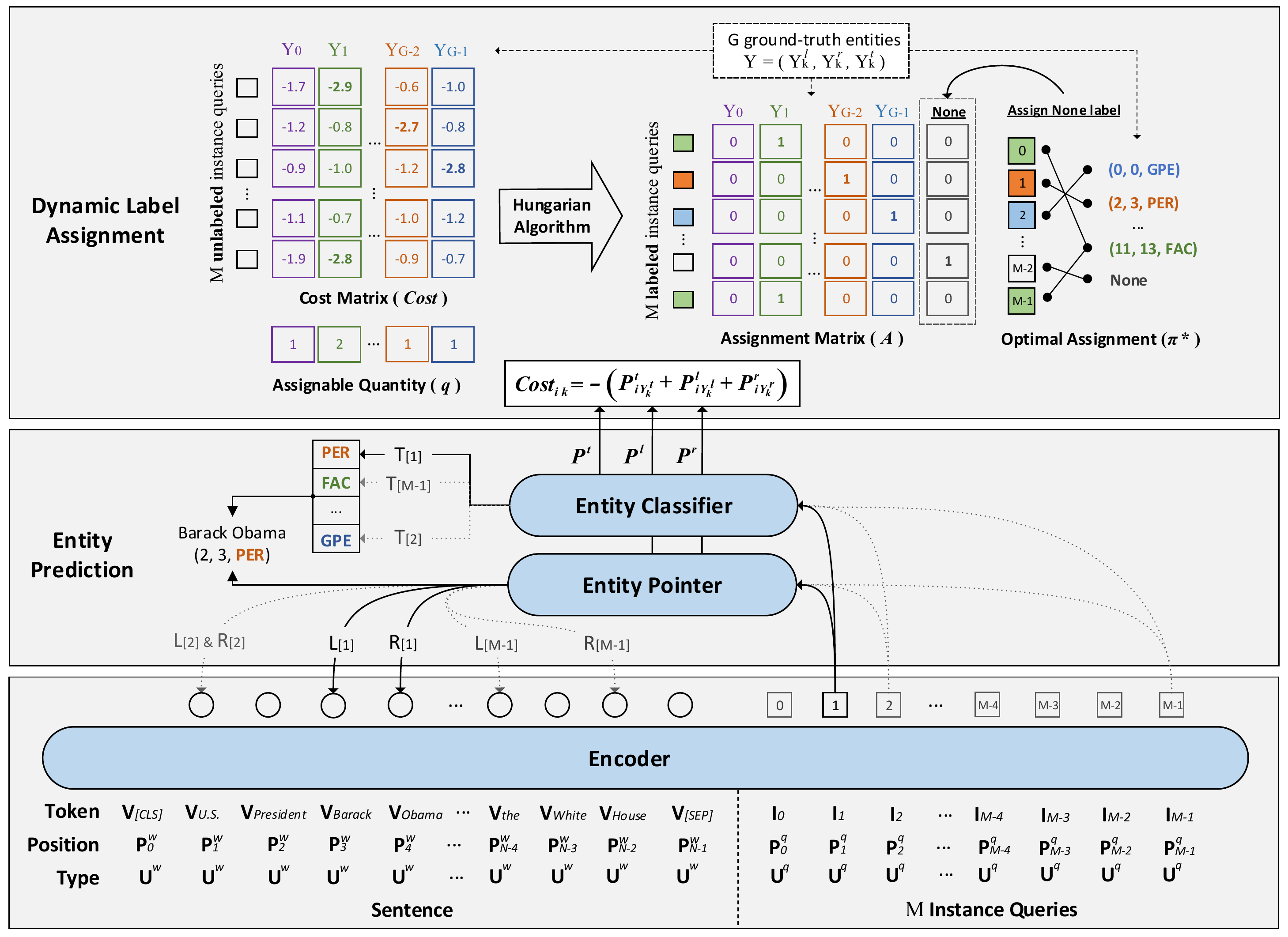}
  \caption{The overall architecture of the model. }
  \label{fig:overview}
\end{figure*}

In this section, we first introduce the task formulation in \cref{3.1}, and then describe our method. As shown in Figure \ref{fig:overview}, our method consists of three components: the Encoder (\cref{3.2}), the Entity Prediction (\cref{prediction}) and the Dynamic Label Assignment (\cref{label}). The encoder encodes both the sentence and instance queries. Then for each instance query, we perform entity localization and entity classification using Entity Pointer and Entity Classifier respectively. 
For training the model, we introduce a dynamic label assignment mechanism to assign gold entities to the instance queries in \cref{label}.
 
\subsection{Task Formulation}
\label{3.1}
We use $(X, Y)$ to denote a training sample, where $X$ is a sentence consisting of $N$ words labeled by a set of triples $Y = \{\text{< } Y_k^l, Y_k^r, Y_k^t \text{ >}\}^{G-1}_{k=0}$.
$Y_k^l \in [0, N-1]$, $Y_k^r \in [0, N-1]$ and $Y_k^t\in \mathcal{E}$ are the indices for the left boundary, right boundary and entity type of the $k$-th entity, where $\mathcal{E}$ is a finite set of entity types.
In our approach, We set up $M (M>G)$ global and learnable instance queries $I = \mathbb{R}^{M \times h}$, each of which (denoted as a vector of size $h$) extracts one entity from the sentence.
They are randomly initialized and can learn the query semantics automatically during training.
Thus we define the task as follows: given an input sentence $X$, the aim is to extract the entities $Y$ based on the learnable instance queries $I$.


\subsection{Encoder}
\label{3.2}

Model input consists of two sequences, the sentence $X$ of length $N$ and the instance queries $I$ of length $M$. The encoder concatenates them into one sequence and encodes them simultaneously.

\paragraph{Input Embedding} We calculate the token embeddings $E_{tok} $, position embeddings $E_{pos}$ and type embeddings $E_{typ}$ of the input from two sequences as follows ($E_{tok},E_{pos},E_{typ} \in \mathbb{R}^{(N+M)\times h} $):

\begin{equation}
    \begin{aligned} 
    E_{tok} &= \operatorname{Concat}(V, I) \\
    E_{pos} &= \operatorname{Concat}(P^w, P^q) \\
    E_{typ} &= \operatorname{Concat}([U^w]^N , [U^q]^M)
\end{aligned}
\end{equation}


\noindent where $V\in \mathbb{R}^{N\times h}$ are token embeddings of the word sequence, $I\in \mathbb{R}^{M\times h}$ are the vectors of instance queries, $P^w\in \mathbb{R}^{N\times h}$ and $ P^q\in \mathbb{R}^{M\times h}$ are separate learnable position embeddings. $U^w$ and $U^q$ are type embeddings and $[ \cdot ]^N$ means repeating $N$ times. Then the input can be represented as $H^{0} = E_{tok} + E_{pos} + E_{typ}\in R^{(N+M)\times h} $.


\paragraph{One-Way Self-Attention} 

Normal self-attention would let the sentence interact with all instance queries. In such a way, randomly initialized instance queries can affect the sentence encoding and break the semantics of the sentence.
To keep the sentence semantics isolated from the instance queries,
we replace the self-attention in BERT \citep{devlin-etal-2019-bert} with the one-way version:



\begin{equation}
\operatorname{OW-SA}(H)= \alpha HW_v
\end{equation}


\begin{equation}
\alpha=\operatorname{softmax}\left(\frac{HW_q (HW_k)^{T}}{\sqrt{h}} + \mathcal{M} \right) 
\end{equation}

\noindent where $W_q,W_k,W_v \in \mathbb{R}^{h\times h}$ are parameter matrices and $\mathcal{M}\in \{0, -\inf\}^{(N+M)\times(N+M)}$ is a mask matrix for the attention score where elements in $\mathcal{M}$ set to 0 for kept units and $-\inf$ for removed ones. In our formula, the upper right sub-matrix of $\mathcal{M}$ is a full $-\inf$ matrix of size $(N\times M)$ and other elements are zero, which can prevent the sentence encoding from attending on the instance queries.
In addition, the self-attention among instance queries can model the connections between each other, and then enhance their query semantics.


        
    

After BERT encoding, we further encode the sequence at word-level by two bidirectional LSTM layers and $L$ extra transformer layers. Finally we split $H\in \mathbb{R}^{(N+M)\times h}$ into  two parts: the sentence encoding $H^w\in \mathbb{R}^{N\times h}$ and the instance query encoding $H^q\in \mathbb{R}^{M\times h}$.

\subsection{Entity Prediction}
\label{prediction}

Each instance query can predict one entity from the sentence, and with $M$ instance queries, we can predict at most $M$ entities in parallel. Entity prediction can be viewed as a joint task of boundary prediction and category prediction. We design Entity Pointer and Entity Classifier for them respectively.

\paragraph{Entity Pointer} For the $i$-th instance query $H^q_i$, we first interact the query with each word of the sentence by two linear layers. The fusion representation of the $i$-th instance query and $j$-th word is computed as:

\begin{equation}
    S^\delta_{ij} = \operatorname{ReLU}(H^q_iW^q_\delta+H^w_jW^w_\delta)
\end{equation}

\noindent where $\delta \in\{l, r\}$ denotes the left or right boundary and $W^q_\delta,W^w_\delta\in \mathbb{R}^{h\times h}$ are trainable projection parameters. Then we calculate the probability that the $j$-th word of the sentence is a left or right boundary:

\begin{equation}
    P^\delta_{ij} = \operatorname{sigmoid}(S^\delta_{ij}W_\delta + b_\delta)
\end{equation}

\noindent where $W_\delta\in \mathbb{R}^{h}$ and $b_\delta$ are learnable parameters.








\paragraph{Entity Classifier} Entity boundary information are useful for entity typing.
We use {$P^\delta_i = [P^\delta_{i0}, P^\delta_{i1}, \cdots, P^\delta_{iN-1}], \delta \in \{l, r\}$} to weigh all words and then concatenate them with instance queries. The boundary-aware representation of the $i$-th instance query can be calculated as:
\begin{equation}
    S^t_i = \operatorname{ReLU}\left(\left[H^q_iW^q_t;P^l_iH^w;P^r_iH^w \right]\right)
\end{equation}



\noindent where $W^q_t\in \mathbb{R}^{h\times h}$ is a learnable parameter. 
Then we can get the probability of the entity queried by the $i$-th instance query belonging to category $c$:
\begin{equation}
    P^t_{ic} = \frac{\exp(S^t_iW^c_t  + b_t^c)}{\sum_{c^{\prime}\in \mathcal{E}}\exp(S^t_iW^{c^{\prime}}_t + b_t^{c^{\prime}})}
\end{equation}

\noindent where $W_t^{{c^{\prime}}}\in \mathbb{R}^{h}$ and $b_t^{c^{\prime}}$ are learnable parameters.

Finally, the entity predicted by the $i$-th instance query is $\mathcal{T}_i = \left( \mathcal{T}^l_i, \mathcal{T}^r_i, \mathcal{T}^{t}_i \right)$. $\mathcal{T}^l_i = \mathop{\arg\max}_j(  P^l_{ij})$ and $\mathcal{T}^r_i = \mathop{\arg\max}_j(P^r_{ij})$ are the left and right boundary, $\mathcal{T}^{t}_i = \mathop{\arg\max}_c(P^t_{ic})$ is the entity type. 
We perform entity localization and entity classification on all instance queries to extract entities in parallel.
If multiple instance queries locate the same entity but predict different entity types, we keep only the prediction with the highest classification probability.




\subsection{Dynamic Label Assignment for Training}
\label{label}

\paragraph{Dynamic Label Assignment}

Since instance queries are implicit (not in natural language form), we cannot assign gold entities to them in advance. To tackle this, we dynamically assign labels for the instance queries during training. Specifically, we treat label assignment as a Linear Assignment Problem. Any entity can be assigned to any instance query, incurring some cost that may vary depending on the entity-query assignment. We define the cost of assigning the $k$-th entity ($Y_k = \text{< }Y_k^l, Y_k^r, Y_k^t\text{ >}$) to the $i$-th instance query as:

\begin{equation}
    \textit{Cost}_{ik} = -\left(P^t_{iY^t_k} + P^l_{iY^l_k} + P^r_{iY^r_k}\right)
\end{equation}

\noindent where $Y^t_k$, $Y^l_k$ and $Y^r_k$ denote the indices for the entity type, left boundary and right boundary of the $k$-th entity.
It is required to allocate as many entities as possible by assigning at most one entity to each query and at most one query to each entity, in such a way that the total cost of the assignment is minimized.
However, the one-to-one manner does not fully utilize instance queries, and many instance queries are not assigned to gold entities. Thus
we extend the traditional LAP to one-to-many one, where each entity can be assigned to multiple instance queries. The optimization objective of this one-to-many LAP is defined as:





\begin{equation}
\label{lap}
    \begin{aligned}
&\min \sum_{i=0}^{M-1} \sum_{k=0}^{G-1} A_{i k} \textit{Cost}_{i k} \\
& \begin{array}{ll}
s.t.&\sum_{k} A_{i k}\leq 1 \\
&\sum_{i} A_{i k}=q_k \\
&\forall i, k, A_{i k} \in \{0, 1\} \
\end{array} .
\end{aligned}
\end{equation}

\noindent where $A\in \{0, 1\}^{M\times G}$ is the assignment matrix, $G$ denotes the number of the entities and $A_{ik}$ = 1 indicates the $k$-th entity assigned to the $i$-th instance query.
$q_k$ denotes the assignable quantity of the $k$-th gold entity and $Q=\sum_{k}q_k$ denotes the total assignable quantity for all entities.
In our experiments, the assignable quantities of different entities are balanced. 

We then use the Hungarian \citep{kuhn1955hungarian} algorithm to solve Equation \ref{lap}, which yields the label assignment matrix with the minimum total cost. However, the number of instance queries is greater than the total assignable quantity of entity labels ($M>Q$), so some of them will not be assigned to any entity label. 
We assign \texttt{None} label to them by extending a column for the assignment matrix. The new column vector $a$ is set as follows:

\begin{equation}
a_{i} = 
\begin{cases}
0, & \sum_kA_{ik} = 1 \\
1, &\sum_kA_{ik} = 0\end{cases}
\end{equation}

Based on the new assignment matrix $\hat{A}\in \{0, 1\}^{M \times (G+1)}$, we can further get the labels $\hat{Y} = Y.\operatorname{indexby}(\pi^*)$ for $M$ instance queries,
\noindent where $\pi^* = \mathop{\arg\max}\limits_{dim=1}(\hat{A})$ is the label index vector for instance queries under the optimal assignment.


\paragraph{Training Objective}

We have computed the entity predictions for $M$ instance queries in \cref{prediction} and got their labels $\hat{Y}$ with the minimum total assignment cost in \cref{label}. To train the model, we define boundary loss and classification loss. For left and right boundary prediction, we use binary cross entropy function as a loss:


\begin{equation}
\begin{aligned}
    \mathcal{L}_{b} =& - \sum_{\delta \in \{l, r\}} \sum^{M-1}_{i=0} \sum_{j=0}^{N-1}  \mathds{1}[\hat{Y}^\delta_i=j] \log P^\delta_{ij} \\
     &+\mathds{1}[\hat{Y}^\delta_i \neq j] \log \left(1- P^\delta_{ij}\right)
    \end{aligned}
\end{equation}

\noindent and for entity classification we use cross entropy function as a loss:


\begin{equation}
    \mathcal{L}_{t}= -\sum^{M-1}_{i=0} \sum_{c\in \mathcal{E}} \mathds{1}[\hat{Y}^t_i=c]\log P_{ic}^t
\end{equation}

\noindent where $\mathds{1}[\omega]$ denotes indicator function that takes 1 when $\omega$ is true and 0 otherwise.

Follow \citet{article} and \citet{10.1007/978-3-030-58452-8_13}, we add Entity Pointer and Entity Classifier after each word-level transformer layer, and we can get the two losses at each layer. Thus, the total loss on the train set $D$ can be defined as:

\begin{equation}
    \mathcal{L}= \sum_D\sum_{\tau=1}^{L} \mathcal{L}^\tau_{t} + \mathcal{L}^\tau_{b} 
\end{equation}

\noindent where $\mathcal{L}^{\tau}_{t}, \mathcal{L}^{\tau}_{b}$ are classification loss and boundary loss at the $\tau$-th layer. For prediction, we just perform entity prediction at the final layer. 







\begin{table}[]
\centering
\small
\begin{tabular}{lccc}

\toprule
\multirow{2}{*}{Model}   & \multicolumn{3}{c}{ACE04}  \\
 \cmidrule(lr){2-4} 
& Pr.  & Rec. & F1  \\
\midrule

\citet{li-etal-2020-unified} & 85.05 & 86.32 &  85.98 \\
\citet{wang-etal-2020-pyramid}       & 86.08  & 86.48  & 86.28     \\
\citet{yu-etal-2020-named}       & 87.30  & 86.00  & 86.70     \\
\citet{yan2021bartner}  & 87.27  & 86.41  & 86.84     \\
\citet{yang2022} &  86.60  & 87.28 &  86.94 \\
\citet{tan2021sequencetoset}  & 88.46 &  86.10 &  87.26    \\
\citet{shen2021locateandlabel} & 87.44  & 87.38  & 87.41  \\
\midrule
PIQN & \textbf{88.48} & \textbf{87.81}  & \textbf{88.14}  \\
\toprule
\multirow{2}{*}{Model}   & \multicolumn{3}{c}{ACE05}  \\
 \cmidrule(lr){2-4} 
& Pr.  & Rec. & F1  \\
\midrule

\citet{lin-etal-2019-sequence}       & 76.20  & 73.60 & 74.90  \\
\citet{luo-zhao-2020-bipartite}       & 75.00  & 75.20  & 75.10    \\
\citet{li-etal-2021-span}  & - & - &  83.00\\
\citet{wang-etal-2020-pyramid}       & 83.95  & 85.39  & 84.66     \\
\citet{yan2021bartner}  & 83.16  & 86.38  & 84.74     \\
\citet{yu-etal-2020-named}       & 85.20  & 85.60  & 85.40     \\
\citet{yang2022}  & 84.61 & 86.43 & 85.53     \\
\citet{li-etal-2020-unified} & 87.16 & 86.59 & 86.88 \\
\citet{shen2021locateandlabel} & {86.09}   & {87.27} & {86.67}  \\
\citet{tan2021sequencetoset}  & 87.48  & 86.63  & 87.05     \\
\midrule
PIQN    & 86.27  & \textbf{88.60} & \textbf{87.42}  \\

\toprule
\multirow{2}{*}{Model}   & \multicolumn{3}{c}{GENIA}  \\
 \cmidrule(lr){2-4} 
& Pr.  & Rec. & F1  \\
\midrule

\citet{lin-etal-2019-sequence}       & 75.80  & 73.90 & 74.80  \\
\citet{luo-zhao-2020-bipartite}       & 77.40  & 74.60  & 76.00    \\
\citet{wang-etal-2020-hit}       & 78.10  & 74.40  & 76.20  \\
\citet{yang2022}  & 78.08 & 78.26 & 78.16 \\
\citet{li-etal-2020-unified}$\dagger$ & 81.14 & 76.82 & 78.92 \\

\citet{wang-etal-2020-pyramid}       & 79.45  & 78.94  & 79.19     \\
\citet{yan2021bartner}  & 78.87  & 79.6  & 79.23     \\
\citet{tan2021sequencetoset}  & 82.31  & 78.66  & 80.44     \\
\citet{yu-etal-2020-named}       & 81.80  & 79.30  & 80.50     \\
\citet{shen2021locateandlabel} & 80.19 & 80.89  & 80.54  \\
\midrule
PIQN    & \textbf{83.24} & 80.35 & \textbf{81.77} \\
\toprule
\multirow{2}{*}{Model}   & \multicolumn{3}{c}{KBP17}  \\
 \cmidrule(lr){2-4} 
& Pr.  & Rec. & F1  \\
\midrule

\citet{DBLP:conf/tac/JiPZNMMC17}       & 76.20 & 73.00 & 72.80  \\
\citet{lin-etal-2019-sequence}       & 77.70  & 71.80 & 74.60  \\
\citet{luo-zhao-2020-bipartite}       & 77.10 & 74.30 & 75.60  \\
\citet{li-etal-2020-unified} & 80.97  & 81.12  & 80.97  \\
\citet{tan2021sequencetoset}  & 84.91  & 83.04  & 83.96     \\
\citet{shen2021locateandlabel} & {85.46}  & {82.67}  & {84.05}  \\
\midrule
PIQN & \textbf{85.67}  & \textbf{83.37} & \textbf{84.50} \\
\toprule



\multirow{2}{*}{Model}   & \multicolumn{3}{c}{NNE}  \\
 \cmidrule(lr){2-4} 
& Pr.  & Rec. & F1  \\
\midrule

\citet{li-etal-2020-unified}$\ddagger$ & 53.13  & 56.67  & 54.84  \\
\citet{wang-lu-2018-neural}       & 77.40 & 70.10 & 73.60  \\
\citet{ringland-etal-2019-nne}       & 91.80 & 91.00 & 91.40  \\

\citet{tan2021sequencetoset}$\ddagger$  & 93.01 & 89.21 & 91.07     \\
\citet{shen2021locateandlabel}$\ddagger$ & 92.86 & 91.12 & 91.98  \\

\citet{wang-etal-2020-pyramid}$\dagger$       & 92.64  & 93.53  & 93.08     \\
\midrule
PIQN & \textbf{93.85} & \textbf{94.23} & \textbf{94.04} \\
\bottomrule

\end{tabular}
\caption{Results for \textit{\textbf{nested}} NER task. $\dagger$ means the reproduction on the same preprocessed dataset and $\ddagger$ means that we run the code on the unreported dataset.}
\label{tab:nested}
\end{table}







\section{Experiment Settings}
\subsection{Datasets}


To provide empirical evidence for the effectiveness of the proposed model, we conduct our experiments on eight English datasets, including five nested NER datasets: ACE04 \citep{doddington-etal-2004-automatic} , ACE05  \citep{2005-automatic}, KBP17 \citep{DBLP:conf/tac/JiPZNMMC17}, GENIA \citep{10.5555/1289189.1289260}, NNE\citep{ringland-etal-2019-nne} and three flat NER dataset: FewNERD \citep{ding-etal-2021-nerd}, CoNLL03 \citep{tjong-kim-sang-de-meulder-2003-introduction}, OntoNotes \citep{pradhan-etal-2013-towards}, and one Chinese flat NER dataset: MSRA \citep{levow-2006-third}.
FewNERD and NNE are two datasets with large entity type inventories, containing 66 and 114 fine-grained entity types.
Please refer to Appendix \ref{app:statistic} for statistical information about the datasets.






\subsection{Implementation Details}

In our experiments, we use pretrained BERT \citep{devlin-etal-2019-bert} in our encoder. For a fair comparison, we use \texttt{bert-large} on ACE04, ACE05, NNE, CoNLL03 and OntoNotes, \texttt{bert-base} on KBP17 and FewNERD, \texttt{biobert-large} \citep{chiu-etal-2016-train} on GENIA and \texttt{chinese-bert-wwm} \citep{cui-etal-2020-revisiting} on Chinese MSRA. For all datasets, we train our model for 30-60 epochs and use the Adam Optimizer \citep{adam} with a linear warmup-decay learning rate schedule.
We initialize all instance queries using the normal distribution $\mathcal{N}(0.0, 0.02)$. See Appendix \ref{app:id} for more detailed parameter settings and Appendix \ref{app:baseline} for all baseline models.

\subsection{Evaluation Metrics}


We use strict evaluation metrics that an entity is confirmed correct when the entity boundary and the entity type are correct simultaneously. We employ precision, recall and F1-score to evaluate the performance. We also report the F1-scores on the entity localization and entity classification subtasks in \cref{ab} and Appendix \ref{app:subtask}. We consider the localization as correct when the left and right boundaries are predicted correctly. Based on the accurately localized entities, we then evaluate the performance of entity classification.

\section{Results and Analysis}

\subsection{Performance}

\paragraph{Overall Performance} Table \ref{tab:nested} illustrates the performance of the proposed model as well as baselines on the nested NER datasets. We observe significant performance boosts on the nested NER datasets over previous state-of-the-art models, achieving F1-scores of 81.77\%, 88.14\%, 87.42\% and 84.50\% on GENIA, ACE04, ACE05, KBP17 and NNE datasets with +1.23\%, +0.73\%, +0.37\%, +0.45\% and +0.96\% improvements.
Our model can be applied to flat NER. As shown in Table \ref{tab:flat}, our model achieves state-of-the-art performance on the FewNERD and Chinese MSRA datasets with +1.44\% and +0.88\% improvements. On the CoNLL03 and OntoNotes datasets, our model also achieves comparable results.
Compared with the type-specific query-based method \citep{li-etal-2020-unified}, our model improves by +2.85\%, +2.16\%, +0.54\%, +3.53\% on the GENIA, ACE04, ACE05 and KBP17 datasets. We believe there are three reasons: (1) Rather than relying on external knowledge to inject semantics, instance queries can learn query semantics adaptively, avoiding the sensitivity to hand-constructed queries of varying quality. (2) Each query no longer predicts a group of entities of a specific type, but only one entity. 
This manner refines the query to the entity level with more precise query semantics.
(3) Instance queries are fed into the model in parallel for encoding and prediction, and different instance queries can exploit the intrinsic connections between entities.

\begin{table}[!h]
\centering
\small
\begin{tabular}{lccc}

\toprule
\multirow{2}{*}{Model}   & \multicolumn{3}{c}{FewNERD}  \\
 \cmidrule(lr){2-4} 
& Pr.  & Rec. & F1  \\
\midrule

\citet{ding-etal-2021-nerd}  & 65.56 & 68.78 & 67.13 \\
\citet{shen2021locateandlabel}$\ddagger$  & 64.69    &    70.87     &   67.64  \\
\citet{tan2021sequencetoset}$\ddagger$  &  67.37  & 69.12 & 68.23    \\
\midrule
PIQN    & \textbf{70.16} & \textbf{69.18} & \textbf{69.67}  \\

\toprule
\multirow{2}{*}{Model}   & \multicolumn{3}{c}{English CoNLL03}  \\
 \cmidrule(lr){2-4} 
& Pr.  & Rec. & F1  \\
\midrule

\citet{peters-etal-2018-deep} & -  & -  & 92.22 \\
\citet{devlin-etal-2019-bert} & -  & -  & 92.80 \\
\citet{li-etal-2020-unified}$\ast$ & 92.47 & 93.27 & 92.87 \\
\citet{yu-etal-2020-named}$\ast$ & 92.85 &  92.15 & 92.50     \\
\citet{shen2021locateandlabel} & 92.13  & {93.73}  & 92.94  \\
\midrule
PIQN    & \textbf{93.29}  & 92.46  & 92.87  \\
\toprule
\multirow{2}{*}{Model}   & \multicolumn{3}{c}{English OntoNotes}  \\
 \cmidrule(lr){2-4} 
& Pr.  & Rec. & F1  \\
\midrule

\citet{li-etal-2020-unified}$\ast$ & 91.34 & 88.39 & 89.84 \\
\citet{yu-etal-2020-named}$\ast$ & 89.74 &  89.92 & 89.83     \\
\citet{yan2021bartner} & 89.99 &  90.77  & 90.38\\
\citet{xu-etal-2021-better} & 90.14  & 91.58  & 90.85     \\
\midrule
PIQN    &  \textbf{91.43}  & 90.73  & \textbf{90.96}  \\

\toprule
\multirow{2}{*}{Model}   & \multicolumn{3}{c}{Chinese MSRA}  \\
 \cmidrule(lr){2-4} 
& Pr.  & Rec. & F1  \\
\midrule

\citet{devlin-etal-2019-bert} & - & - & 92.60 \\
\citet{li-etal-2020-unified}$\dagger$ & 90.38 & 89.00 & 89.68     \\
\citet{shen2021locateandlabel}$\ddagger$ &  92.20 & 90.72 & 91.46\\
\citet{tan2021sequencetoset}$\ddagger$ & 93.21 & 91.97 & 92.58    \\
\midrule
PIQN    &  \textbf{93.61}  & \textbf{93.35}  & \textbf{93.48}  \\

\bottomrule
\end{tabular}
\caption{Results for \textit{\textbf{flat}} NER task. $\ast$ means the result reproduced by \cite{yan2021bartner}$, \dagger$ means the reproduction on the same preprocessed dataset and $\ddagger$ means that we run the code on the unreported dataset.}
\label{tab:flat}
\end{table}



\begin{table*}[]
\centering
\small
\begin{tabular}{lcccccccccc}
\toprule
\multirow{2}{*}{Model}   & \multicolumn{5}{c}{ACE04} & \multicolumn{5}{c}{GENIA}  \\

 \cmidrule(lr){2-6}  \cmidrule(lr){7-11} 
& Loc. F1 & Cls. F1 & Pr.  & Rec. & F1 & Loc. F1 & Cls. F1 & Pr.  & Rec. & F1  \\
\midrule
Default & \textbf{92.23} & \textbf{91.53}   &   \textbf{88.48} & \textbf{87.81}  & \textbf{88.14}    &  \textbf{84.43} &  \textbf{87.83} & \textbf{83.24} & \textbf{80.35} & \textbf{81.77}\\
\midrule
w/o Dynamic LA & 88.22 & 88.29 & 80.95 & 83.99 & 82.43 & 77.01 & 81.90 &  73.56 & 72.30 & 72.93 \\
w/o OvM LA   & 89.22 & 87.61 & 87.04 & 81.68 & 84.28 &  83.87 &  87.38 &  83.02 &	79.57 &	81.26 \\
w/o One Way SA &  91.90 &  90.62 &  87.56 & 86.75 & 87.16 &  84.11 &  87.21 & 82.94  & 79.53  & 81.20 \\
w/o Query Interaction   & 91.84 & 90.42 &  88.21       & 86.26    &    87.22 &  83.87 &  87.05 &  83.15 &	79.15 &	81.10 \\
\bottomrule
\end{tabular}
\caption{Ablation Study. (1) \textbf{w/o Dynamic LA}: replace dynamic label assignment to static label assignment, i.e., assign labels to instance queries in the order of the entities' occurrence in the sentence. (2) \textbf{w/o OvM LA}: replace the one-to-many label assignment to one-to-one, i.e., set the number of queries to which each entity can be assigned to be 1. (3) \textbf{w/o One Way SA}: encode sentences and instance queries using the original BERT. (4) \textbf{w/o Query Interaction}: eliminate interactions between instance queries by masking the attention weights between them.}
\label{tab:ablation}
\end{table*}

\paragraph{Inference Speed}

We compare the inference speed on ACE04 and NNE, as shown in Table \ref{tab:speed}. Compared to the type-specific query method \citep{li-etal-2020-unified}, our model not only improves the performance, but also gains significant inference speedup. In particular, on the NNE dataset with 114 entity types, our model speeds up by 30.46$\times$ and improves performance by +39.2\%. This is because \citet{li-etal-2020-unified} requires one inference for each type-specific query, while our approach performs parallel inference for all instance queries and only needs to be run once. 
We also compare previous state-of-the-art models \citep{tan2021sequencetoset, shen2021locateandlabel} and our method is still faster and performs better.

\begin{table}[]
\centering
\small
\begin{tabular}{p{2.4cm}>{\centering\arraybackslash}p{0.75cm}>{\centering\arraybackslash}p{0.75cm}>{\centering\arraybackslash}p{0.75cm}>{\centering\arraybackslash}p{0.75cm}}
\toprule
\multirow{2}{*}{Model}   & \multicolumn{2}{>{\centering\arraybackslash}p{1.5cm}}{ACE04}& \multicolumn{2}{>{\centering\arraybackslash}p{1.5cm}} {NNE} \\
 \cmidrule(lr){2-3} \cmidrule(lr){4-5} 
& Speedup & F1 & Speedup & F1 \\
\midrule
\citet{li-etal-2020-unified}  & 1.00$\times$ & 85.98 & 1.00$\times$ & 54.84\\ 
\citet{tan2021sequencetoset}   & 1.40$\times$ & 87.26 & 22.18$\times$ & 91.07\\ 
\citet{shen2021locateandlabel} & 0.96$\times$ & 87.41 & 11.41$\times$ & 91.98\\ 
\midrule
PIQN &  \textbf{2.16}$\boldsymbol{\times}$ & \textbf{88.14} & \textbf{30.46}$\times$ & \textbf{94.04}\\ 

\bottomrule
\end{tabular}
\caption{Inference Speed on ACE04 and NNE. All experiments are conducted on a single NVIDIA RTX A6000 Graphical Card with 48G graphical memory.}
\label{tab:speed}
\end{table}





\subsection{Ablation Study}
\label{ab}







In this section, we analyze the effects of different components in PIQN.
As shown in Table \ref{tab:ablation}, we have the following observations: (1) Compared to the static label assignment in order of occurrence, the dynamic label assignment shows significant improvement on localization, classification, and NER F1-score, which improves NER F1-score by +5.71\% on ACE04 and +8.84\% on GENIA. This shows that modeling label assignment as a LAP problem enables dynamic assignment of optimal labels to instance queries during training, eliminating the incorrect bias when pre-specifying labels. Furthermore, one-to-many for label assignment is more effective than one-to-one, improving the F1-score by +3.86\% on ACE04 and +0.51\% on GENIA. 
(2) The one-way self-attention blocks the attention of sentence encoding on instance queries, which improves the F1-score by +0.98\% on ACE04 and +0.57\% on GENIA. It illustrates the importance of keeping the semantics of the sentence independent of the query. In contrast, semantic interactions between queries are effective, which improves the F1-score by +0.92\% on ACE04 and +0.67\% on GENIA. 
The major reason is that entities in the same sentence are closely related and the interaction between instance queries can capture the relation between them.

\subsection{Analysis}

\label{analysis1}

In order to analyze the query semantics learned by the instance query in the training, we randomly selected several instance queries and analyzed the locations and types of entities they predicted.


\begin{figure}[h]
  \centering
  \includegraphics[width=0.9\linewidth]{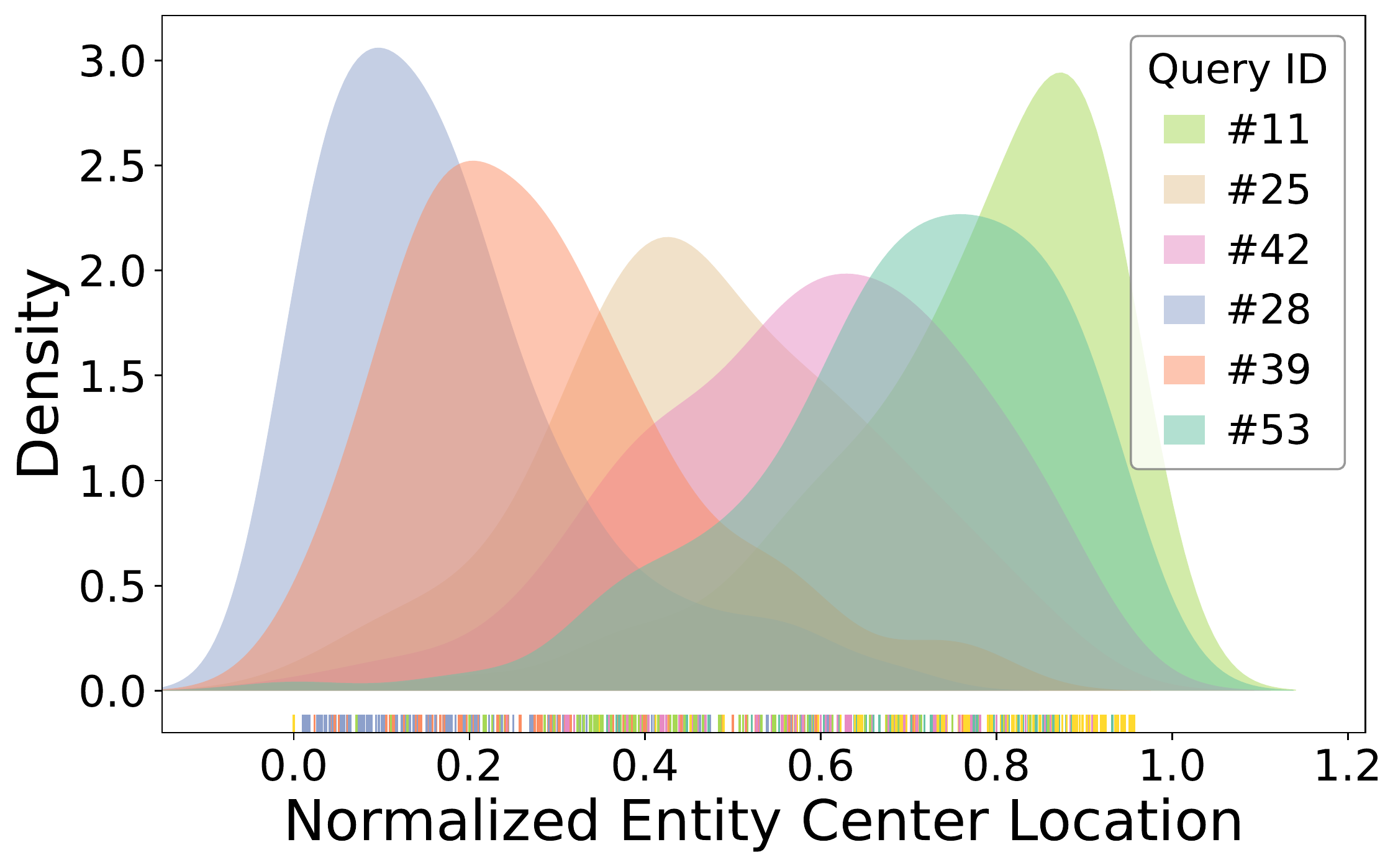}
  \caption{Kernel density estimation of entity distribution at different locations.}
  \label{fig:position}
\end{figure}

\paragraph{Entity Location} We normalize the predicted central locations of the entities and use kernel density estimation to draw the distribution of the predicted entity locations for different queries, as shown in Figure \ref{fig:position}. We observe that different instance queries focus on entities at different positions, which means that the instance queries can learn the query semantics related to entity position.
For example, instance queries \#28 and \#39 prefer to predict entities at the beginning of sentences, while \#11 and \#53 prefer entities at the end.

\begin{figure}[h]
  \centering
  \includegraphics[width=0.9\linewidth]{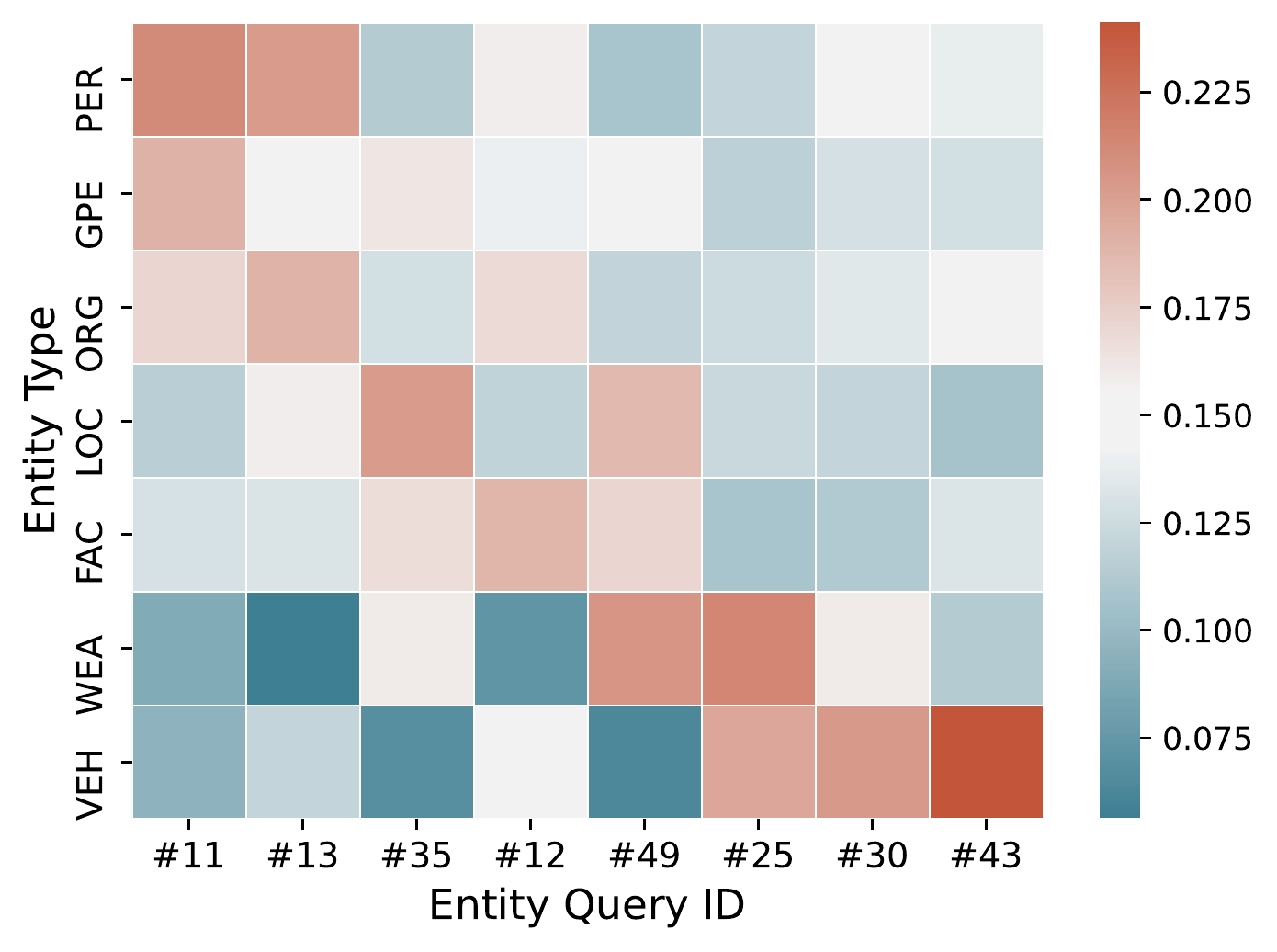}
  \caption{Co-occurrence statistics between instance queries and different entity types}
  \label{fig:type}
\end{figure}

\paragraph{Entity Type} We count the co-occurrence of different instance queries and different entity types they predicted. To eliminate the imbalance of entity types, we normalize the co-occurrence matrix on the entity type axis. As shown in Figure \ref{fig:type},  different instance queries have preferences for different entity types. For example, instance queries \#11 and \#13 prefer to predict \texttt{PER} entities, \#30 and \#43 prefer \texttt{VEH} entities, \#25 and \#49 prefer \texttt{WEA} entities, \#12 prefers \texttt{FAC} entities, and \#35 prefers \texttt{LOC} entities.



We also analyze the auxiliary loss, the dynamic label assignment mechanism, and the performance on entity localization and classification, please see the Appendix \ref{app:analysis}.

\begin{table*}[!]
\small
\centering
  \begin{tabular}{m{0.15cm}m{7cm}m{7.5cm}}
  \toprule
  \# & Sentence with Gold Entities & Prediction $\leftarrow$ Instance Query IDs \\
  \midrule

1 & {\color{Cyan4}[\textsuperscript{0}}A number of powerful international companies and commercial agencies , such as {\color{Cyan4}[\textsuperscript{12}}Ito Bureau of {\color{blue}[\textsuperscript{15}}Japan{\color{blue}\textsuperscript{15}]$_{\text{GPE}}$}{\color{Cyan4}\textsuperscript{15}]$_{\text{ORG}}$} , {\color{Cyan4}[\textsuperscript{17}}Han Hua Group of {\color{blue}[\textsuperscript{21}}South Korea{\color{blue}\textsuperscript{22}]$_{\text{GPE}}$}{\color{Cyan4}\textsuperscript{22}]$_{\text{ORG}}$} , {\color{Cyan4}[\textsuperscript{24}}Jeffrey Group of {\color{blue}[\textsuperscript{27}}the US{\color{blue}\textsuperscript{28}]$_{\text{GPE}}$}{\color{Cyan4}\textsuperscript{28}]$_{\text{ORG}}$} , {\color{Cyan4}[\textsuperscript{30}}etc{\color{Cyan4}\textsuperscript{30}]$_{\text{ORG}}$}{\color{Cyan4}\textsuperscript{30}]$_{\text{ORG}}$} . participated in this Urumchi Negotiation Meeting . & 
\makecell[l]{
\cmark\ \ {\color{Cyan4}(24, 28, ORG)} $\leftarrow$ 0 23 33 45 51  \\
\cmark\ \ {\color{blue}(27, 28, GPE)} $\leftarrow$ 2 3 19 26 27 46 50  \\
\cmark\ \ {\color{blue}(15, 15, GPE)} $\leftarrow$ 9 11 14 42  \\
\cmark$\quad \quad \quad \quad $ ··· $\quad \quad \quad \quad $ ···\\
\cmark\ \ {\color{Cyan4}(0, 30, ORG)} $\leftarrow$ 10 20 24 37 53 55  \\
\xmark\ \ {\color{Cyan4}(12, 30, ORG)} $\leftarrow$ 16 22 47 57  \\
\textbf{None} $\leftarrow$ 1 12 13 15 17 21 29 30 31 32 34 35 40 49 52 59  \\
}
\\ \midrule

2 & For example , as instant messaging migrates to cell phones or hand - held computer organizers , {\color{red}[\textsuperscript{17}}consumers{\color{red}\textsuperscript{17}]$_{\text{PER}}$} won ' t want to have to install multiple services on these devices , said {\color{red}[\textsuperscript{33}}Brian Park{\color{red}\textsuperscript{34}]$_{\text{PER}}$} , {\color{red}[\textsuperscript{36}}senior product for {\color{Cyan4}[\textsuperscript{39}}Yahoo !{\color{Cyan4}\textsuperscript{40}]$_{\text{ORG}}$} Communications Services{\color{red}\textsuperscript{42}]$_{\text{PER}}$} . & 
\makecell[l]{
\xmark\ \ {\color{Cyan4}(39, 42, ORG)} $\leftarrow$ 0 2 15 19 26 27 29 35 46 49 50  \\
\cmark\ \ {\color{red}(17, 17, PER)} $\leftarrow$ 1 10 20 22 24 32 37 47 53 55 57  \\
\cmark\ \ {\color{red}(33, 34, PER)} $\leftarrow$ 6 9 11 12 14 18 34 38 42 48 59  \\
\cmark\ \ {\color{red}(36, 42, PER)} $\leftarrow$ 8 17 25 28 30 31 36 40 54 56 58  \\
\textbf{None} $\leftarrow$ 3 4 5 7 13 16 21 23 33 39 41 43 44 45 51 52  \\
}
\\\midrule

3 & {\color{red}[\textsuperscript{0}}Hector Rodriguez{\color{red}\textsuperscript{1}]$_{\text{PER}}$} told the hearing of {\color{Cyan4}[\textsuperscript{6}}the Venezuelan consumer protection agency{\color{Cyan4}\textsuperscript{10}]$_{\text{ORG}}$} that {\color{Cyan4}[\textsuperscript{12}}Bridgeton Firestone{\color{Cyan4}\textsuperscript{13}]$_{\text{ORG}}$} knew about the tyre defects for many months and should be held responsible for the accidents . &

\makecell[l]{
\cmark\ \ {\color{red}(0, 1, PER)} $\leftarrow$ 1 10 20 24 32 37 47 53 55  \\
\cmark\ \ {\color{Cyan4}(12, 13, ORG)} $\leftarrow$ 2 3 19 26 27 35 46 49 50  \\
\xmark\ \ {\color{red}(7, 8, PER)} $\leftarrow$ 4 7 12 18 38 39 41 43 44  \\
\cmark\ \ {\color{Cyan4}(6, 10, ORG)} $\leftarrow$ 5 6 9 11 14 21 48 57 59  \\
\xmark\ \ {\color{blue}(7, 7, GPE)} $\leftarrow$ 8 25 28 30 31 36 40 54 56 58  \\
\textbf{None} $\leftarrow$ 0 13 15 16 17 22 23 29 33 34 42 45 51 52  \\
}

\\

\bottomrule
 \end{tabular}
 \caption{Cases Study. In the left column, the label in the lower right corner indicates the type of entity, and the superscripts indicate the positions of the left and right boundary words. In the right column, we show the correspondence between the instance queries and the predicted entities.}
  \label{tab:case}
\end{table*}

\section{Case Study}

\label{cs}

Table \ref{tab:case} shows a case study about model predictions. Our model can recognize nested entities and long entities well. In case 1, the entities of length 31 or with the three-level nested structure are predicted accurately. And thanks to the one-to-many dynamic label assignment mechanism, each entity can be predicted by multiple instance queries, which guarantees a high coverage of entity prediction. However, the model's ability to understand sentences is still insufficient, mainly in the following ways: (1) There is a deficiency in the understanding of special phrases. \textit{Yahoo ! Communications Services} in case 2 is misclassified as \texttt{ORG}, but in fact \textit{Yahoo !} is \texttt{ORG}. (2) Over-focus on local semantics. In case 3, the model misclassifies \textit{Venezuelan consumer} as \texttt{PER}, ignoring the full semantics of the long phrase \textit{the Venezuelan consumer protection agency}, which should be \texttt{ORG}. (3) Insensitivity to morphological variation. The model confused \textit{Venezuelan} and \textit{Venezuela}, and misidentified the former as \texttt{GPE} in case 3.


\section{Conclusion}

In this paper, we propose Parallel Instance Query Network for nested NER, where a collection of instance queries are fed into the model simultaneously and can predict all entities in parallel. The instance queries can automatically learn query semantics related to entity types or entity locations during training, avoiding manual constructions that rely on external knowledge. To train the model, we design a dynamic label assignment mechanism to assign gold entities for these instance queries.
Experiments on both nested and flat NER datasets demonstrate that the proposed model achieves state-of-the-art performance.

\section*{Acknowledgments}

This work is supported by the Key Research and Development Program of Zhejiang Province, China (No. 2021C01013), the National Key Research and Development Project of China (No. 2018AAA0101900), the Chinese Knowledge Center of Engineering Science and Technology (CKCEST) and MOE Engineering Research Center of Digital Library.

\bibliographystyle{acl_natbib}
\bibliography{custom}

\clearpage
\newpage

\appendix

\section{Datasets}


\label{app:statistic}

\paragraph{GENIA} \citep{10.5555/1289189.1289260} is an English biology nested named entity dataset and contains 5 entity types, including \texttt{DNA}, \texttt{RNA}, \texttt{protein}, \texttt{cell line}, and \texttt{cell type} categories. Follow \citet{yu-etal-2020-named}, we use 90\%/10\% train/test split and evaluate the model on the last epoch.

\paragraph{ACE04 and ACE05} \citep{doddington-etal-2004-automatic, 2005-automatic} are two English nested datasets, each of them contains 7 entity categories. We follow the same setup as previous work \citet{katiyar-cardie-2018-nested, lin-etal-2019-sequence}.

\paragraph{KBP17} \citep{DBLP:conf/tac/JiPZNMMC17} has 5 entity categories, including \texttt{GPE}, \texttt{ORG}, \texttt{PER}, \texttt{LOC}, and \texttt{FAC}. We follow \citet{lin-etal-2019-sequence} to split all documents into 866/20/167 documents for train/dev/test set.

\paragraph{NNE} \citep{ringland-etal-2019-nne} is a English nested NER dataset with 114 fine-grained entity types. Follow \citet{wang-etal-2020-pyramid}, we keep the original dataset split and pre-processing.

\paragraph{FewNERD} \citep{ding-etal-2021-nerd} is a large-scale English flat NER dataset with 66 fine-grained entity types. Follow \citet{ding-etal-2021-nerd}, we adopt a standard supervised setting. 


\paragraph{CoNLL03}  \citep{tjong-kim-sang-de-meulder-2003-introduction} is an English dataset with 4 types of named entities: \texttt{LOC}, \texttt{ORG}, \texttt{PER} and \texttt{MISC}. Follow \citet{yan2021bartner, yu-etal-2020-named}, we train our model on the train and development sets.

\paragraph{OntoNotes} \citep{pradhan-etal-2013-towards} is an English dataset with 18 types of named entity, consisting of 11 types and 7 values. We use the same train, development, test splits as \citet{li-etal-2020-unified}.

\paragraph{Chinese MSRA} \citep{levow-2006-third} is a Chinese dataset with 3 named entity types, including \texttt{ORG}, \texttt{PER}, \texttt{LOC}. We keep the original dataset split and pre-processing.


In Table \ref{tab:statistics} and Table \ref{tab:statistics2}, we report the number of sentences, the number of sentences containing nested entities, the average sentence length, the total number of entities, the number of nested entities, the nesting ratio, the maximum and the average number of entities in a sentence on all datasets.

\label{app:nestedner}



\begin{table*}[!ht]
\centering
\small
\begin{tabular}{>{\centering\arraybackslash}p{0.6cm}>{\centering\arraybackslash}p{0.6cm}>{\centering\arraybackslash}>{\centering\arraybackslash}p{0.6cm}>{\centering\arraybackslash}p{0.6cm}>{\centering\arraybackslash}p{0.6cm}>{\centering\arraybackslash}p{0.6cm}>{\centering\arraybackslash}p{0.6cm}>{\centering\arraybackslash}p{0.6cm}>{\centering\arraybackslash}p{0.6cm}>{\centering\arraybackslash}p{0.6cm}>{\centering\arraybackslash}p{0.6cm}>{\centering\arraybackslash}p{0.6cm}>{\centering\arraybackslash}p{0.7cm}>{\centering\arraybackslash}p{0.7cm}>{\centering\arraybackslash}p{0.7cm}}
\toprule
\multirow{2}{*}{}   & \multicolumn{3}{c}{ACE04}& \multicolumn{3}{c}{ACE05} & \multicolumn{3}{c}{KBP17} & \multicolumn{2}{c}{GENIA} &
\multicolumn{3}{c}{NNE}\\
 \cmidrule(lr){2-4}  \cmidrule(lr){5-7} \cmidrule(lr){8-10} \cmidrule(lr){11-12}  
 \cmidrule(lr){13-15}  
& Train  & Dev & Test & Train  & Dev & Test & Train  & Dev & Test & Train   & Test & Train  & Dev & Test  \\
\midrule
\#S  &  6200 &  745 &  812 &  7194 &  969 &  1047 &  10546 &  545 & 4267 &  16692 &   1854 & 43457  & 1989  & 3762 \\
\#NS  &  2712 &  294 &  388 &  2691 &  338 &  320 &  2809 &  182 &  1223 &  3522 &   446 & 28606 & 1292 & 2489 \\
\#E &  22204 &  2514 &  3035 &  24441 &  3200 &  2993 &  31236 &  1879 &  12601 &  50509 &    5506  & 248136 & 10463 & 21196 \\
\#NE &  10149 &  1092 & 1417  & 9389 &  1112 &  1118 &  8773 &  605 &  3707 &  9064 &    1199  & 206618 & 8487 & 17670\\
NR &  45.71 & 46.69 &  45.61 & 38.41 & 34.75 &  37.35 &  28.09 & 32.20 &  29.42 &  17.95 &    21.78 & 83.27 & 81.11 & 83.36\\
AL &  22.50 &  23.02 &  23.05 &  19.21 &  18.93 &  17.2 &  19.62 &  20.61 &  19.26 &  25.35 &    25.99 & 23.84 & 24.20 & 23.80 \\
\#ME &  28 & 22 &  20 & 27 & 23 &  17 &  58 & 15 &  21 &  25 &  14 & 149 & 58 &  64 \\
\#AE &  3.58 & 3.37 &  3.73 & 3.39 & 3.30 &  2.86 &  2.96 & 3.45 &  2.95 &  3.03 & 2.97  & 5.71 & 5.26 & 5.63\\
\bottomrule

\end{tabular}
\caption{Statistics of the \textbf{\textit{nested}} datasets used in the experiments. \#S: the number of sentences, \#NS: the number of sentences containing nested entities, \#E: the total number of entities, \#NE: the number of nested entities, NR: the nesting ratio (\%), AL: the average sentence length, \#ME: the maximum number of entities in a sentence, \#AE: the average number of entities in a sentence}
\label{tab:statistics}
\end{table*}

\begin{table*}[!ht]
\centering
\small
\begin{tabular}{>{\centering\arraybackslash}p{0.6cm}>{\centering\arraybackslash}p{0.6cm}>{\centering\arraybackslash}>{\centering\arraybackslash}p{0.6cm}>{\centering\arraybackslash}p{0.6cm}>{\centering\arraybackslash}p{0.6cm}>{\centering\arraybackslash}p{0.6cm}>{\centering\arraybackslash}p{0.6cm}>{\centering\arraybackslash}p{0.6cm}>{\centering\arraybackslash}p{0.6cm}>{\centering\arraybackslash}p{0.6cm}>{\centering\arraybackslash}p{0.6cm}>{\centering\arraybackslash}p{0.6cm}>{\centering\arraybackslash}p{0.7cm}>{\centering\arraybackslash}p{0.7cm}>{\centering\arraybackslash}p{0.7cm}}
\toprule
\multirow{2}{*}{}   & \multicolumn{3}{c}{CoNLL03}& \multicolumn{3}{c}{OntoNotes} & 
\multicolumn{3}{c}{FewNERD} & \multicolumn{3}{c}{Chinese MSRA}\\
 \cmidrule(lr){2-4}  \cmidrule(lr){5-7} \cmidrule(lr){8-10}  \cmidrule(lr){11-13}
& Train  & Dev & Test & Train  & Dev & Test & Train  & Dev & Test &  Train  & Dev & Test  \\
\midrule
\#S  &  14041 &  3250 &  3453 &  49706 &  13900 &  10348 & 131965  & 18824  & 37648 & 41728 & 4636 & 4365 \\
\#E &  23499 &  5942 &  5648 &  128738 &  20354 &  12586   & 340247 & 48770 & 96902 & 70446 & 4257 & 6181 \\
AL &  14.50 &  15.80 &  13.45 & 24.94 &  20.11 &  19.74  & 24.49 & 24.61 & 24.47 & 46.87 & 46.17 & 39.54 \\
\#ME &  20 & 20 &  31 & 32 & 71 &  21 & 50 & 35 &  49 & 125 & 18 & 461 \\
\#AE &  1.67 & 1.83 &  1.64 & 2.59 & 1.46 &  1.22   & 2.58 & 2.59 & 2.57 & 1.69 & 0.92 & 1.42\\
\bottomrule

\end{tabular}
\caption{Statistics of the \textbf{\textit{flat}} datasets used in the experiments. \#S: the number of sentences, \#E: the total number of entities, AL: the average sentence length, \#ME: the maximum number of entities in a sentence, \#AE: the average number of entities in a sentence}
\label{tab:statistics2}
\end{table*}

\section{Implementation Details}

\label{app:id}

In default setting, we set the number of instance queries $M=60$, and the total assignable quantity $Q= M \times 0.75=45$. 
To ensure that the assignable quantities of different entities are balanced, we randomly divide $Q$ to different entities and adjust each division to be larger than $Q/G$, where $G$ is the number of the ground-truth entities.
When the number of entities is more than the total assignable quantity, we specify $Q=G$. We have also tried other configurations that will be discussed in Appendix \ref{comb}.
We set $L$ word-level transformer layers after BERT and set auxiliary losses in each layer. In the default setting $L$ equals 5.
We compare the effect of different auxiliary layers on the model performance, which will be discussed in Appendix \ref{aux}. Since the instance queries are randomly initialized and do not have query semantics at the initial stage of training, we first fix the parameters of BERT and train the model for 5 epochs, allowing the instance queries to initially learn the query semantics. When decoding entities, we filter out the predictions with localization probability and classification probability less than the threshold 0.6 and 0.8, respectively.

\section{Baselines}
\label{app:baseline}
We compare PIQN with the following baselines:

\begin{itemize}
    \item \textbf{ARN} \citep{lin-etal-2019-sequence} designs a sequence-to-nuggets architecture for nested mention detection, which first identifies anchor words and then recognizes the mention boundaries.
    \item \textbf{HIT} \citep{wang-etal-2020-hit} designs a head-tail detector and a  token interaction tagger, which can leverage
the head-tail pair and token interaction to express the nested structure.
    \item \textbf{Pyramid} \citep{wang-etal-2020-pyramid} presents a layered neural model for nested entity recognition, consisting of a stack of inter-connected layers.
    \item \textbf{Biaffine} \citep{yu-etal-2020-named} formulates NER as a structured prediction task and adopts a dependency parsing approach for NER.
    \item \textbf{BiFlaG} \citep{luo-zhao-2020-bipartite} designs a bipartite flat-graph network with two subgraph modules for outermost and inner entities.
    \item \textbf{BERT-MRC} \citep{li-etal-2020-unified} formulates the NER task as a question answering task. They construct type-specific queries using semantic prior information for entity categories.
    \item \textbf{BARTNER} \citep{yan2021bartner} formulates NER as an entity span sequence generation problem and uses a unified Seq2Seq model with the pointer mechanism to tackle flat, nested, and discontinuous NER tasks.
    \item \textbf{Seq2Set} \citep{tan2021sequencetoset} formulates NER as an entity set prediction task. Different from \citet{strakova-etal-2019-neural}, they utilize a non-autoregressive decoder to predict entity set.
    \item \textbf{Locate\&Label} \citep{shen2021locateandlabel} treats NER as a joint task of boundary regression and span classification and proposed a two-stage identifier of locating entities first and labeling them later.
\end{itemize}

For a fair comparison, we did not compare with \citet{Sun_Wang_Li_Feng_Tian_Wu_Wang_2020, li-etal-2020-flat, NEURIPS2019_452bf208} on Chinese MSRA because they either used glyphs or an external lexicon or a larger pre-trained language model. In addition, some works \citep{wang-etal-2021-improving, wang2022damonlp} used search engines to retrieve input-related contexts to introduce external information, and we did not compare with them as well.

\section{Analysis}
\label{app:analysis}

\subsection{Analysis of Auxiliary Loss}
\label{aux}
Many works \citep{article,10.1007/978-3-030-58452-8_13} have demonstrated that the auxiliary loss in the middle layer introduces supervised signals in advance and can improve model performance.
We compared the effect of the different number of auxiliary-loss layers on the model performance (F1-score on ACE04). Overall, the model performs better as the number of auxiliary-loss layers increases. The model achieves the best results when the number of layers equals 5.

\begin{figure}[h]
  \centering
  \includegraphics[width=0.8\linewidth]{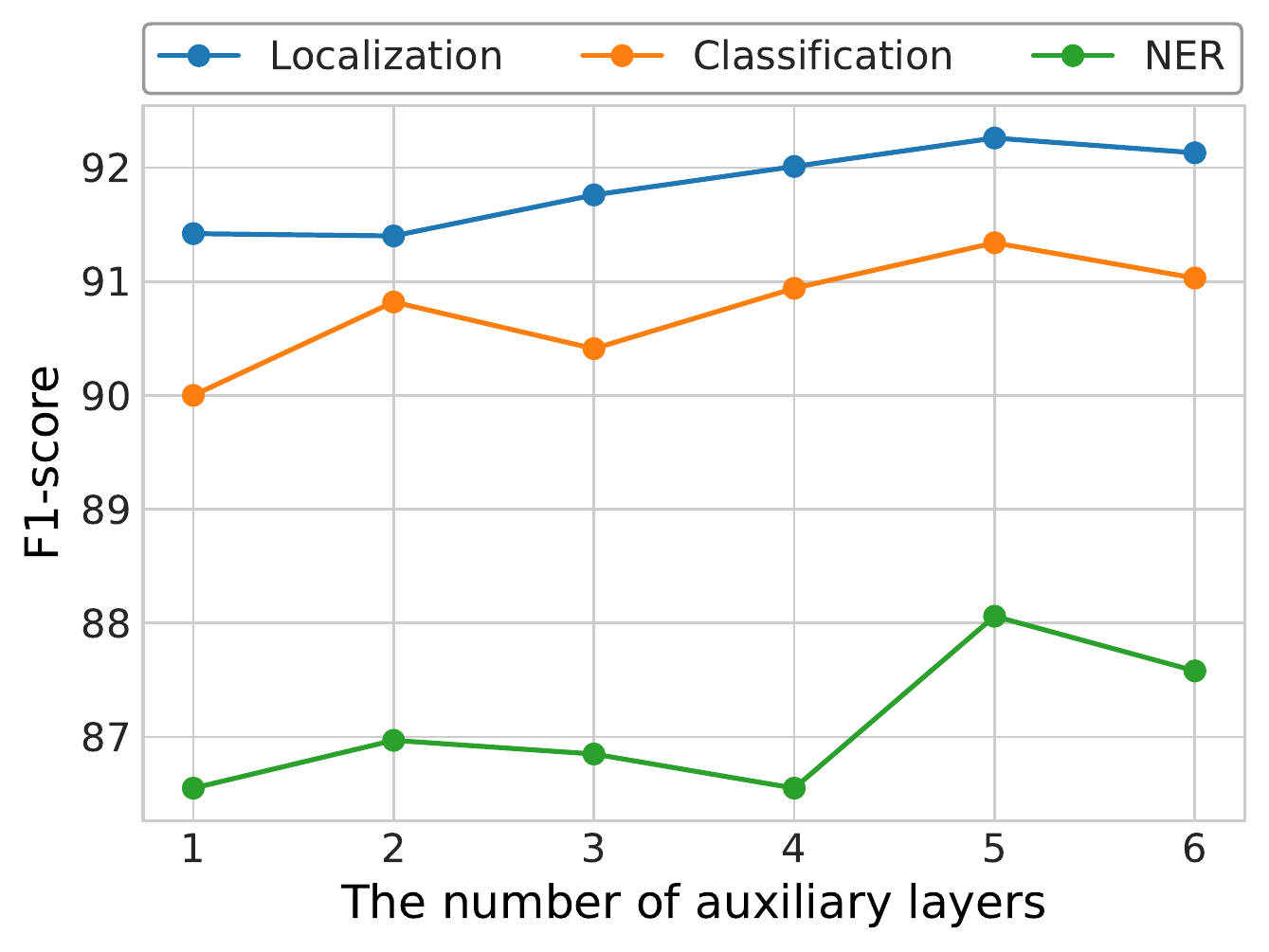}
  \caption{Analysis of Auxiliary Loss}
  \label{fig:auxiliary_layers}
\end{figure}

\subsection{Analysis of Two Subtasks}

\label{app:subtask}

We compare the model performance on entity localization and entity classification subtasks on the ACE04 dataset, as shown in Table \ref{tab:clsandloc}. Compared with the previous state-of-the-art models \citep{tan2021sequencetoset, shen2021locateandlabel}, our model achieves better performance on both entity localization and entity classification subtasks. This illustrates that the instance queries can automatically learn their query semantics about location and type of entities, which is consistent with our analysis in \ref{analysis1}.

\begin{table}[h]
\centering
\small
\begin{tabular}{lccc}

\toprule
\multirow{2}{*}{Model}   & \multicolumn{3}{c}{Localization}  \\
 \cmidrule(lr){2-4} 
& Pr.  & Rec. & F1  \\
\midrule
 \citet{tan2021sequencetoset} & 92.75  &      90.24    &    91.48    \\
\citet{shen2021locateandlabel} & 92.28  &      90.97   &     91.62  \\
PIQN    & {92.56}      &  \textbf{91.89}    &    \textbf{92.23}  \\

\bottomrule
\toprule
\multirow{2}{*}{Model}   & \multicolumn{3}{c}{Classification}  \\
 \cmidrule(lr){2-4} 
& Pr.  & Rec. & F1  \\
\midrule
\citet{tan2021sequencetoset}  & 95.36   &     86.03      &  90.46    \\
\citet{shen2021locateandlabel} & 95.40  &      86.75   &     90.87  \\
PIQN    &  \textbf{95.59}    &     \textbf{87.81}     &    \textbf{91.53} \\

\bottomrule
\end{tabular}
\caption{Localization and Classification Performance on ACE04}
\label{tab:clsandloc}
\end{table}

\subsection{Analysis of Label Assignment}

\label{comb}

\begin{table}[h]
\centering
\small
\begin{tabular}{lccccc}
\toprule
$(M, Q)$ & Loc. F1 & Cls. F1 & Pr.  & Rec. & F1  \\
\midrule

(60, 15) & 91.05 & 90.15 & 87.57 & 85.67 & 86.61   \\
(60, 30) & 91.76 & 90.37 & 88.23 & 86.16& 87.18  \\
(60, 45) & \textbf{92.23} & \textbf{91.53}   &  \textbf{88.48} & \textbf{87.81}  & \textbf{88.14} \\
(60, 50) & 92.01 & 90.81 & 87.38 & 87.12 & 87.25  \\
\midrule
(30, 15)  & 91.26 & 89.66 & 88.61 & 84.88 & 86.70    \\
(60, 30) & 91.76 & 90.37 & 88.23 & 86.16& 87.18  \\
(90, 45) & \textbf{91.88} & \textbf{90.56} & \textbf{88.23} & \textbf{86.46} & \textbf{87.34}   \\
(120, 60) & 91.75 & 90.45  & 87.19 & 86.56 & 86.87  \\


\bottomrule
\end{tabular}
\caption{Analysis on Dynamic Label Assignment for different combinations of the number $M$ of instance queries and the total assignable quantity $Q$ of labels.}
\label{tab:com}
\end{table}

We analyze the impact of dynamic label assignment on model performance for different combinations of the number $M$ of instance queries and the total assignable quantity $Q$ of labels. 
From Table \ref{tab:com}, we observe that (1) there is a tradeoff between $M$ and $Q$, and the model achieves the best performance with a ratio of 4:3. With this setting, the ratio of positive to negative instances of instance queries is 3:1.
(2) The number of instance queries and the total assignable quantity is not as large as possible, and an excessive number may degrade the model performance. In our experiments $(M,Q) = (60, 45)$ is the best combination.

\label{comb}

\end{document}